\documentclass{article} 
\usepackage{iclr2026_conference}
\iclrfinalcopy

\usepackage{microtype}
\usepackage{hyperref}
\usepackage{url}
\usepackage{booktabs}
\usepackage{amsmath}
\usepackage{amssymb}
\usepackage{graphicx}
\usepackage{multirow}
\usepackage{xcolor}
\usepackage{algorithm}
\usepackage{algorithmic}
\usepackage{wrapfig}
\usepackage{caption}
\usepackage{pgfplots}
\usepackage{pifont}
\pgfplotsset{compat=1.18}
\usetikzlibrary{decorations.pathreplacing,calligraphy,patterns}

\definecolor{darkblue}{rgb}{0, 0, 0.5}
\hypersetup{colorlinks=true, citecolor=darkblue, linkcolor=darkblue, urlcolor=darkblue}
\newcommand{\good}[1]{\textcolor{teal}{#1}}
\newcommand{\bad}[1]{\textcolor{red}{#1}}
\newcommand{\up}[1]{\textcolor{teal}{\scriptsize{(+#1)}}}
\newcommand{\down}[1]{\textcolor{red}{\scriptsize{(-#1)}}}

\newcommand{\ours}{\textsc{IntentScore}}

\title{IntentScore: Intent-Conditioned Action Evaluation for Computer-Use Agents}

\author{Rongqian Chen \quad Yu Li \quad Zeyu Fang \quad Sizhe Tang \quad Weidong Cao \quad Tian Lan \\
\\
George Washington University, Washington D.C., USA \\
\texttt{\{rongqianc, yul, joey.fang, s.tang1, weidong.cao, tlan\}@gwu.edu}
}

\begin{document}

\maketitle

\begin{abstract}
Computer-Use Agents (CUAs) leverage large language models to execute GUI operations on desktop environments, yet they generate actions without evaluating action quality, leading to irreversible errors that cascade through subsequent steps. We propose \ours{}, a plan-aware reward model that learns to score candidate actions from 398K offline GUI interaction steps spanning three operating systems. \ours{} trains with two complementary objectives: contrastive alignment for state-action relevance and margin ranking for action correctness. Architecturally, it embeds each candidate's planning intent in the action encoder, enabling discrimination between candidates with similar actions but different rationales. \ours{} achieves 97.5\% pairwise discrimination accuracy on held-out evaluation. Deployed as a re-ranker for Agent S3 on OSWorld, an environment entirely unseen during training, \ours{} improves task success rate by 6.9 points, demonstrating that reward estimation learned from heterogeneous offline trajectories generalizes to unseen agents and task distributions.
\end{abstract}
\section{Introduction}
\label{sec:intro}
Computer-Use Agents (CUAs) leverage large language models (LLMs) to plan and execute graphical user interface (GUI) operations on desktop environments. 
At each step, the agent observes the current screen (typically as a screenshot), reasons about the task progress, and generates executable GUI actions such as mouse clicks, keyboard inputs, and scrolls \cite{liu2025scalecua, wang2025opencua, agashe2025agents1, agashe2025agent}.
Recent systems such as \cite{gonzalez2025unreasonable, yang2025gta1} have demonstrated impressive capabilities on complex multi-step tasks spanning web browsing, document editing, and system administration.

However, these systems share a critical limitation: a single LLM call produces a single action or simply selects the first or most frequent one, discarding the opportunity to compare alternatives~\cite{sager2026comprehensive, renze2024self}.
Recent studies show that CUAs frequently produce errors due to misalignment with the task context or intent~\cite{sager2026comprehensive}.
These errors may cascade since the environment state shifts and recovery is difficult: once a wrong button is clicked or a dialog is dismissed, the environment state changes and recovery is difficult \cite{lee2026agentic}. A natural mitigation is to sample multiple candidate actions, evaluate, rank and select the best one~\cite{tang2026agent}.
However, using the same LLM to self-evaluate its own candidates doubles inference cost and introduces self-preference bias~\cite{wang2024large,madaan2023self}, while methods that require parallel environment instances for test-time scaling are impractical under real-world compute and latency constraints. What is needed is a lightweight, independently trained model that can reliably distinguish correct actions from plausible-but-wrong alternatives without additional LLM calls or environment copies.

We observe that large-scale offline GUI trajectory datasets already provide the raw material to build such a model. Datasets like AgentNet~\citep{wang2025opencua} contain hundreds of thousands of interaction steps across multiple operating systems, each annotated with per-step correctness labels. Crucially, different operating systems (Windows, Mac, Ubuntu) instantiate distinct MDPs that share abstract GUI patterns (dialog flows, menu hierarchies, form layouts) yet differ in visual appearance and interaction conventions. A reward model trained on this multi-environment data should learn patterns of action correctness that transfer across environment boundaries~\citep{ishfaq2024offline, xia2025agentrm}. We formalize this intuition through the lens of in-context reinforcement learning (ICRL)~\citep{moeini2025survey}: rather than learning a policy, we learn a \emph{reward estimator} from offline trajectories spanning a diverse family of MDPs, which generalizes to unseen environments and agents at deployment time without further training.


We propose \ours{}, a lightweight plan-aware dual-encoder reward model that instantiates this framework. \ours{} is pretrained on 398K GUI interaction steps across three operating systems and finetuned on the target Ubuntu domain, learning two complementary objectives: \emph{contrastive alignment} (InfoNCE), which captures state-action relevance, and \emph{reward learning} (margin ranking), which captures action correctness. A key finding is that contrastive alignment alone assigns nearly identical scores to correct and incorrect actions, and reward learning is essential to inject the correctness signal. Architecturally, \ours{} embeds each candidate's planning \emph{intent} in the action encoder rather than the state encoder, enabling discrimination between candidates with similar surface actions but different rationales. At deployment, \ours{} serves as a reward-guided re-ranker: it scores each candidate action generated by the CUA backbone and selects the highest-quality one, requiring no additional LLM calls, environment copies, or agent modifications.

Our contributions are:
\begin{itemize}
    \item A plan-aware dual-encoder reward model that embeds each candidate's planning intent in the action encoder, enabling discrimination between candidates with similar surface actions but different rationales.
    \item A dual-objective training framework combining contrastive alignment (InfoNCE) and reward learning (margin ranking), addressing the finding that alignment alone captures state-action relevance but not action correctness (score gap 0.005 vs.\ 0.046).
    \item A two-stage training pipeline (cross-OS pretraining on 398K steps across three OS families followed by target-domain finetuning) that yields +2.6 points on Ubuntu evaluation through cross-environment transfer.
    \item End-to-end deployment validation: \ours{} integrated as a re-ranker for CUA baseline on OSWorld achieves 52.1\% task success rate (+6.9 points), demonstrating that reward estimation learned from offline multi-environment data transfers to an unseen agent and task distribution.
\end{itemize}
\section{Related work}
\label{sec:related}

\paragraph{Computer-Use Agents.}
LLM-based GUI automation has progressed rapidly along two axes: \emph{architecture and data scaling}, and \emph{action selection quality}. On the first axis, systems have improved by scaling training data~\citep{cheng2024seeclick, wang2024mobile}, introducing large-scale annotated trajectory datasets~\citep{wang2025opencua}, and designing specialized agent-computer interfaces~\citep{yang2024swe}. Skill-based augmentation~\citep{cuaskill} and self-evolving agents that learn from experience~\citep{sun2025seagent} further extend agent capabilities. On benchmarks such as OSWorld~\citep{osworld}, WebArena~\citep{zhou2024webarena}, and AndroidWorld~\citep{rawlesandroidworld}, end-to-end success rates remain well below human performance, with failures frequently cascading from local action errors~\citep{zheng2024gpt, sager2026comprehensive}.

On the second axis, recent work has begun to address action selection quality. Agent S3~\citep{gonzalez2025unreasonable} samples multiple candidates and selects via Best-of-N. GTA-1~\citep{yang2025gta1} applies test-time scaling by sampling multiple action proposals and using an MLLM judge to select the best one, while Agent Alpha~\citep{tang2026agent} introduces tree search over candidate trajectories. WMA Web Agent~\citep{chae2024world} takes a world-model approach, predicting the next state for each candidate via a learned transition model before scoring with a value function. Computer-Using World Model~\citep{guan2026computer} further explores this direction by employing a diffusion model to simulate future screen states, enabling action evaluation through predicted visual outcomes. These methods improve action selection but rely on heavyweight components: additional LLM judge calls (GTA-1), environment rollouts (Agent Alpha), or a separate transition/diffusion model (WMA, CU-World Model). \ours{} instead trains a single lightweight reward model (13M parameters) that directly estimates action quality from (state, action, intent), requiring no additional LLM calls, environment copies, or transition model, while generalizing across environment boundaries via multi-OS offline training.

\paragraph{In-context reinforcement learning.}
In-context reinforcement learning (ICRL) studies how agents can internalize the ability to adapt to unseen environments from offline trajectories, enabling zero-shot generalization without per-task gradient updates~\citep{moeini2025survey, lin2024transformers}. Algorithm Distillation~\citep{laskincontext} distills RL learning histories into a Transformer's forward pass, while DPT~\citep{lee2023supervised} pretrains on multi-task offline datasets to learn in-context decision-making. Subsequent work extends ICRL to partially observable settings~\citep{sondistilling} and broader task distributions~\citep{polubarovvintix}. A common thread is that training across diverse MDP families yields capabilities that transfer to unseen environments, precisely the property we exploit for GUI action evaluation. However, prior ICRL work focuses on \emph{policy learning} in low-dimensional control tasks. \ours{} instead addresses \emph{reward estimation} in a high-dimensional, real-world setting with visual observations, natural language plans, and executable code, learning from multi-OS GUI trajectories to score actions in unseen environments and agents at deployment time.

\section{Method}
\label{sec:method}

\subsection{Problem Formulation}
\label{sec:ProblemFormulation}

Computer-Use Agents interact with desktop environments to complete user-specified tasks.
We formalize each interaction as a Markov decision process $M = (\mathcal{S}, \mathcal{A}, T, R)$, where $\mathcal{S}$ is the state space, $\mathcal{A}$ is the action space, $T(s_{t+1} \mid s_t, a_t)$ is the transition function, and $R(s_t, a_t) \to \{0,1\}$ is the reward function indicating per-step correctness.
At each time step $t$, the agent observes a state $s_t = (v_t, c_t)$ consisting of a screenshot $v_t$ and textual context $c_t = (\text{obs}_t, \text{inst}, \text{reflect}_t)$, where $\text{obs}_t$ is the current observation, $\text{inst}$ is the task instruction, and $\text{reflect}_t$ is a reflection on the previous action.
The agent then selects a GUI action $a_t \in \mathcal{A}$ specified by executable code and screen coordinates.
Different operating systems instantiate different MDPs, as Windows, Mac, and Ubuntu share abstract GUI patterns yet differ in visual layout, widget behavior, and interaction conventions.
We denote a family of $K$ such MDPs as $\mathcal{M} = \{M_k\}_{k=1}^{K}$.
 
At each step, the agent generates $N$ candidate actions via parallel LLM sampling with temperature.
Each candidate $i$ produces an executable action $a_t^{(i)} = (\text{act}_t^{(i)}, \text{code}_t^{(i)}, \text{xy}_t^{(i)})$, where $\text{act}_t^{(i)}$ is a natural-language action description, $\text{code}_t^{(i)}$ is the executable code, and $\text{xy}_t^{(i)}$ are the screen coordinates.
Alongside the action, each candidate generates a planning intent $g_t^{(i)} = \text{thought}_t^{(i)}$, the LLM's rationale for why the action should be taken.
Standard practice selects the first or most frequent candidate and discards the rest.
If we can instead evaluate the quality of each candidate, we can select better actions without retraining the LLM, achieving inference-time scaling.

\paragraph{From offline trajectories to in-context reward learning.}
Large-scale GUI trajectory datasets such as AgentNet \citep{wang2025opencua} provide the raw material to build such an evaluator.
AgentNet contains trajectories across multiple operating systems, each annotated with per-step correctness labels:
\begin{equation}
    \mathcal{D} = \bigcup_{k=1}^{K} \left\{ \tau_j^{(k)} = \big(s_1^{(j)}, a_1^{(j)}, r_1^{(j)}, \ldots, s_{T_j}^{(j)}, a_{T_j}^{(j)}, r_{T_j}^{(j)}\big) \right\}_{j=1}^{N_k}
    \label{eq:dataset}
\end{equation}
where $k$ indexes the operating system, $j$ indexes the task within OS $k$, $N_k$ is the number of trajectories in OS $k$, $T_j$ is the trajectory length, and $r_t \in \{0,1\}$ is the per-step correctness annotation.
The multi-OS structure of $\mathcal{D}$ aligns naturally with the in-context reinforcement learning (ICRL) framework \citep{laskincontext, lee2023supervised}, which demonstrates that training on offline trajectories from a diverse family of MDPs yields generalizable capabilities that transfer to unseen environments without per-task fine-tuning.
Theoretical results support such transfer: \citet{zhang2026lisfc} show that Q-value estimation error between two MDPs $M$ and $M'$ is bounded by their structural distance $d(M, M')$.
GUI environments across operating systems exhibit small inter-MDP distance, as dialog flows, menu hierarchies, and form layouts are largely shared despite surface-level visual differences.
The dataset $\mathcal{D}$ spanning three OS families therefore serves as a multi-MDP training distribution from which we learn a transferable reward estimator.

Our goal is to learn an intent-conditioned action-value estimator $\hat{Q}_\theta$ parameterized by $\theta$ that selects the best candidate at each step:
\begin{equation}
    a_t^* = \arg\max_{i} \; \hat{Q}_\theta\!\left(s_t,\; a_t^{(i)} \;\middle|\; \mathcal{H}_t,\; g_t^{(i)}\right)
    \label{eq:selection}
\end{equation}
where $\mathcal{H}_t = \{(s_{t'}, a_{t'})\}_{t'=t-W}^{t-1}$ denotes the in-context experience from the $W$ most recent steps and $g_t^{(i)}$ is the planning intent of candidate $i$.
The conditioning on $g_t^{(i)}$ is central to our approach.
The history $\mathcal{H}_t$ contains demonstrations from heterogeneous contexts with varying GUI layouts and action semantics, and without intent conditioning the estimator treats all demonstrations equally regardless of relevance to the candidate's goal.
Conditioning on $g_t^{(i)}$ recontextualizes the history so that only intent-aligned demonstrations contribute to the value estimate.

\subsection{\ours{} reward model architecture}

\begin{figure}[t]
\begin{center}
\includegraphics[width=\linewidth]{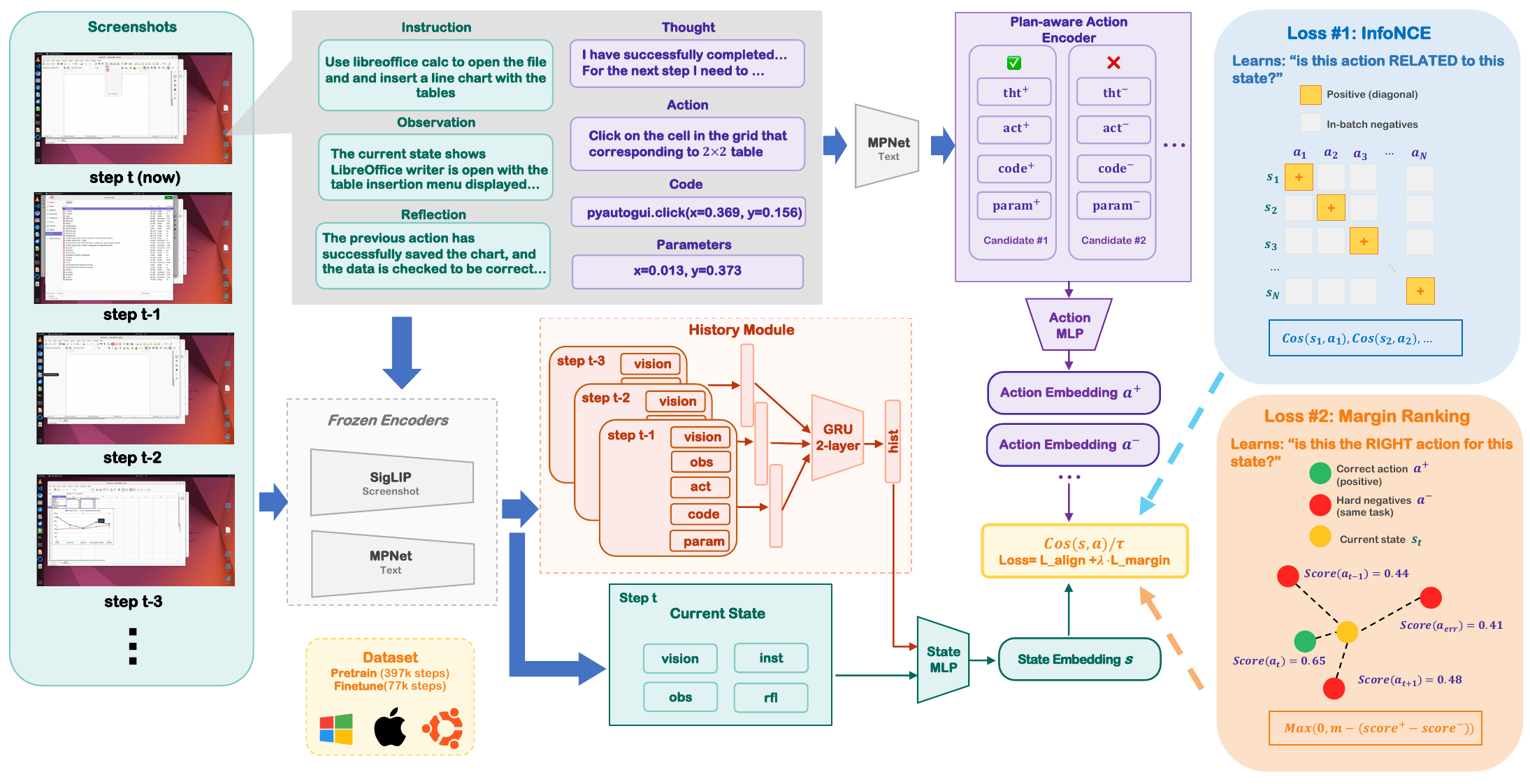}
\end{center}
\caption{Architecture of \ours{}. The state encoder is computed once per step; the intention-aware action encoder is computed per candidate. Reward estimation is temperature-scaled cosine similarity. Training uses a dual objective: state-action alignment (InfoNCE) plus reward learning (margin ranking on hard negatives).}
\label{fig:architecture}
\end{figure}

The reward model consists of three components (Figure~\ref{fig:architecture}). An \textbf{in-context history encoder} summarizes the previous $K{=}3$ steps via a 2-layer GRU~\citep{chung2014empirical} over frozen SigLIP~\citep{siglip} screenshot embeddings and frozen MPNet~\citep{song2020mpnet} text embeddings (observation, action, code, coordinates). A \textbf{state encoder} maps the current screenshot, textual context, and history summary to an L2-normalized 384d vector $\mathbf{s}$, computed once per step and shared across all candidates. An \textbf{intention-aware action encoder} maps each candidate's execution details—action description, code, coordinates—\emph{and its planning intention} to an L2-normalized 384d vector $\mathbf{a}$. The planning intention is deliberately placed in the action encoder rather than the state encoder: because each candidate proposes a different rationale (e.g., ``press Enter to save'' vs.\ ``click Cancel to go back''), this yields distinct action embeddings and enables discrimination between candidates with similar surface actions but different intents. The estimated reward is the temperature-scaled cosine similarity $f(s_t, a_t) = \mathbf{s}^\top \mathbf{a} / \tau$, where $\tau$ is learnable. The full model has ${\sim}$13M trainable parameters and scores each candidate in under 1ms on CPU. Architectural details are in Appendix~\ref{app:architecture}.

\subsection{Dual-objective training: contrastive alignment and reward learning}
\label{sec:training}

\ours{} learns two complementary objectives from offline trajectories. \emph{Contrastive alignment} captures which actions are relevant to which states, i.e., the structural patterns of GUI interaction. \emph{Reward learning} captures which actions are correct, i.e., the per-step reward signal. Each is addressed with a dedicated loss: InfoNCE for alignment, margin ranking for reward learning.

\paragraph{State-action alignment via InfoNCE ($\mathcal{L}_{\text{align}}$).}
The first objective learns state-action co-occurrence patterns across the MDP family. For a batch of $B$ state-action pairs $\{(\mathbf{s}_i, \mathbf{a}_i)\}_{i=1}^{B}$, we use symmetric InfoNCE~\citep{infonce}:
\begin{equation}
    \mathcal{L}_{\text{align}} = \frac{1}{2B} \sum_{i=1}^{B} w_i \left[ -\log \frac{e^{\mathbf{s}_i^\top \mathbf{a}_i / \tau}}{\sum_{j=1}^{B} e^{\mathbf{s}_i^\top \mathbf{a}_j / \tau}} - \log \frac{e^{\mathbf{a}_i^\top \mathbf{s}_i / \tau}}{\sum_{j=1}^{B} e^{\mathbf{a}_i^\top \mathbf{s}_j / \tau}} \right]
\end{equation}
where diagonal entries are positives and all off-diagonal entries serve as in-batch negatives. The weight $w_i$ reflects task completion confidence: $w_i = 1.0$ for steps from completed tasks, $0.3$ for failed tasks, and $0.7$ for tasks with unknown completion status. Only correct steps ($r_t = 1$) are used as positive training samples. With batch size 1024, each positive pair is contrasted against 1023 negatives from potentially different operating systems, encouraging the model to learn cross-environment state-action patterns. However, as we show in Section~\ref{sec:alignment_vs_reward}, this objective alone does not yield a useful reward function: incorrect actions that are semantically related to the current screen receive equally high alignment scores.

\paragraph{Reward learning via margin ranking ($\mathcal{L}_{\text{margin}}$).}
The second objective explicitly learns from the reward signal $r_t$ to discriminate correct from incorrect actions. For each training step $t$ with correct action $\mathbf{a}^+$ ($r_t = 1$), we construct hard negatives $\mathbf{a}^-$ from two sources: (1) temporally adjacent steps ($t{\pm}1$ in the same task), whose actions are semantically similar but temporally incorrect, and (2) steps with $r_t = 0$ (labeled-incorrect actions), representing actual mistakes. The margin ranking loss enforces a minimum separation:
\begin{equation}
    \mathcal{L}_{\text{margin}} = \frac{1}{|\mathcal{N}|} \sum_{(a^+, a^-) \in \mathcal{N}} \max\big(0, \; m - f(s, a^+) + f(s, a^-)\big)
\end{equation}
where $m$ is the target margin, $f(s, a) = \mathbf{s}^\top \mathbf{a} / \tau$ is the reward score, and $\mathcal{N}$ is the set of (positive, negative) pairs constructed for the same state $s$. Contrastive alignment teaches which actions are \emph{relevant} to a state; reward learning teaches which action is \emph{correct}. We verify this complementary distinction empirically in Section~\ref{sec:alignment_vs_reward}.

\paragraph{Combined objective.}
The total training objective combines both components:
\begin{equation}
    \mathcal{L} = \mathcal{L}_{\text{align}} + \lambda \, \mathcal{L}_{\text{margin}}
\end{equation}
where $\lambda$ controls the relative weight of reward learning. This dual objective ensures the reward model learns both \emph{what is relevant} (alignment) and \emph{what is correct} (reward) from the offline trajectories.

\subsection{Two-stage training pipeline}
\label{sec:two_stage}

\paragraph{Stage 1: Cross-OS pretraining.}
The reward model is pretrained on 398,828 steps spanning three MDP families: 77,448 Ubuntu steps plus 321,380 Windows and Mac steps from AgentNet. By training across diverse GUI patterns (Windows start menus, Mac docks, Ubuntu panels, various dialog styles), the model learns reward estimation patterns invariant to specific implementations.

\paragraph{Stage 2: Target-domain finetuning.}
The pretrained model is finetuned on the target Ubuntu subset with a reduced learning rate and increased reward learning weight $\lambda$, preserving cross-OS capability while sharpening reward discrimination for the target domain. Deployment on OSWorld (Section~\ref{sec:deployment}) then tests generalization to entirely unseen tasks and agents.

\paragraph{Training details.}
Both stages use AdamW with cosine annealing, batch size 1024, and gradient clipping at max norm 1.0. Random word dropout (30--50\% of words, 75\% probability per sample) is applied to action and code text throughout training, preventing overfitting to surface-level text patterns and improving robustness to paraphrased descriptions from different LLM backbones. Validation combines in-batch retrieval accuracy (30\%) and pairwise adjacent-step discrimination accuracy (70\%) with early stopping. Total training completes on a single A100 GPU. Full hyperparameters are in Appendix~\ref{app:hyperparams}.

\subsection{Deployment on OSWorld}
\label{sec:gating}

\begin{wrapfigure}{r}{0.5\linewidth}
\vspace{-12pt}
\centering
\includegraphics[width=\linewidth]{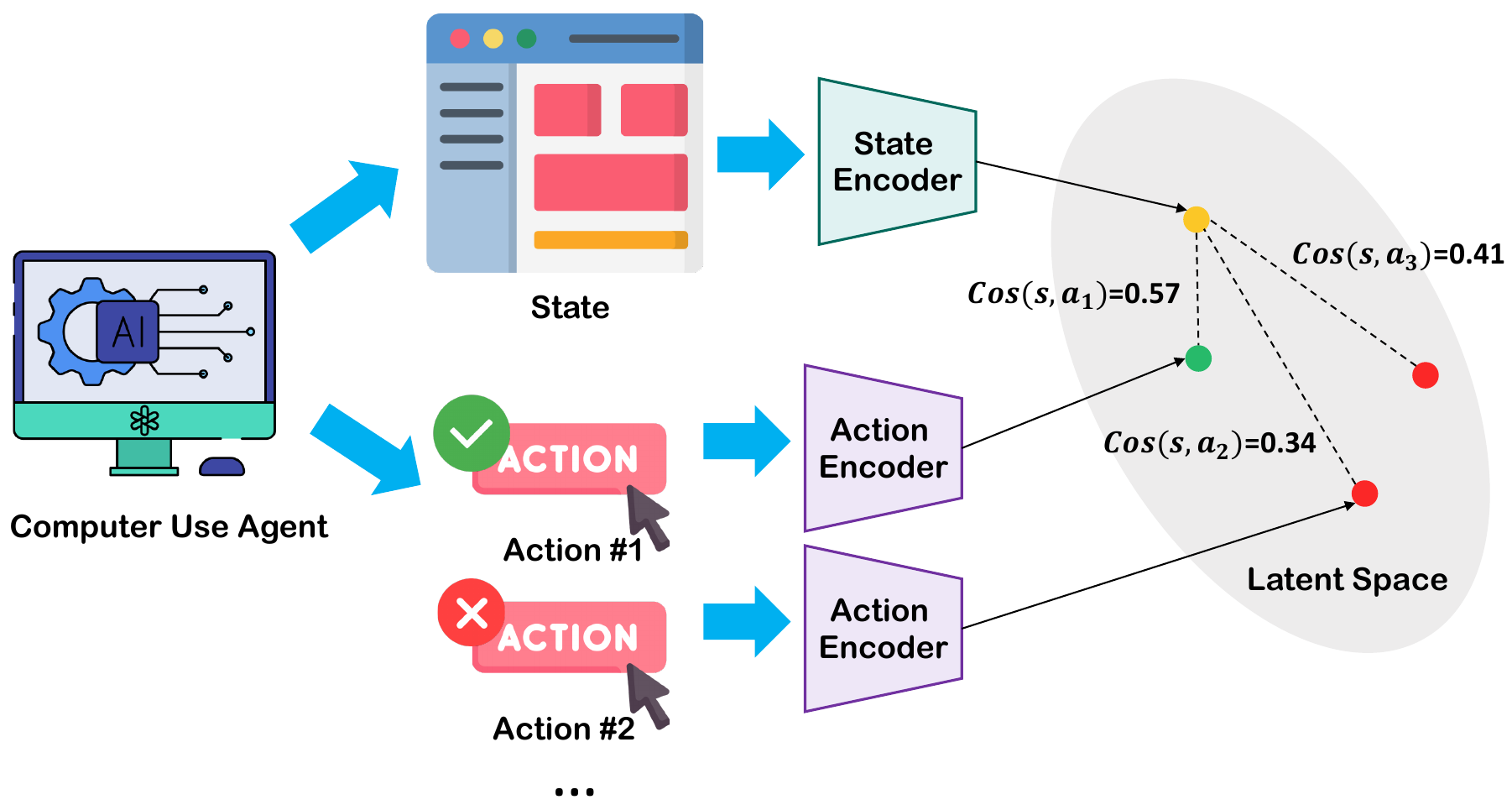}
\vspace{-18pt}
\caption{Deployment inference pipeline. The CUA generates multiple candidate actions for the current state. The state and action encoders map inputs into a shared latent space, where cosine similarity determines action quality.}
\label{fig:inference}
\vspace{-12pt}
\end{wrapfigure}


We deploy \ours{} as a reward-guided re-ranker within Agent S3 on OSWorld, an environment entirely unseen during training. Agent S3 uses GPT-5-mini for planning and UI-TARS-1.5-7B~\citep{uitars} for visual grounding, generating $N$ candidate plans per step via temperature sampling. Candidates whose executable code is identical or whose click coordinates differ by fewer than 20\,px are merged before scoring. \ours{} then scores each unique candidate and selects the highest-scoring one above a threshold $\sigma = 0.10$; if all candidates fall below $\sigma$, the scorer defers to Agent S3's default choice. We restrict to GUI-only actions to isolate the effect of action selection. Field mapping details are in Appendix~\ref{app:field_mapping}.
\section{Results}
\label{sec:results}

\subsection{Experimental setup}
\label{sec:experiments}

\paragraph{Dataset.}
We train on the AgentNet dataset~\citep{wang2025opencua}, which contains 398,828 GUI interaction steps across Ubuntu (77k steps, 5k tasks) and Windows/Mac (321k steps, 17.6k tasks). 
Tasks span four main domains: work, professional, daily, and system, with medium-to-high complexity involving multi-application workflows, professional knowledge requirements, and uncommon feature usage. 
Each step includes a 1920$\times$1080 screenshot, natural-language observation and action descriptions, executable \texttt{pyautogui} code, the agent's thought and reflection, the task instruction, and a correctness label. Data split details are in Appendix~\ref{app:data_splits}.

\paragraph{Deployment benchmark.}
We evaluate end-to-end on OSWorld~\citep{osworld}, a benchmark of 361 real-world computer tasks in fully functional Ubuntu virtual machines. Tasks span 10 application domains: Chrome, GIMP, LibreOffice Calc/Impress/Writer, VLC, VSCode, Thunderbird, OS-level operations, and multi-application workflows. Each task is judged by execution-based scripts that verify the final system state rather than action trajectories, making evaluation robust to alternative solution paths. Crucially, OSWorld is \emph{entirely unseen} during training: neither its tasks, OS images, nor application configurations appear in AgentNet, making it a rigorous test of cross-environment generalization.

\subsection{Offline evaluation on AgentNet}
\label{sec:offline_eval}

All the offline evaluations are pairwise: given a state, the model scores a correct and an incorrect action; the test passes if the correct action receives a higher score. We evaluate on 2,000 held-out test pairs under two settings:
\textbf{Hard}: the negative is drawn from an immediately adjacent step ($t{\pm}1$) in the same task, sharing nearly identical UI context and representing the most realistic challenge;
\textbf{Real Incorrect}: the negative is an action labeled incorrect within the same task, testing detection of actual human-labeled mistakes.
Beyond accuracy, we report the mean pairwise score gap and the fraction of pairs exceeding a 0.10 gap (the deployment override threshold), as these directly predict deployment effectiveness.

\begin{table}[t]
\begin{center}
\small
\caption{Ablation study on training methodology. Hard = adjacent-step discrimination accuracy; Real Inc.\ = detection of labeled-incorrect actions; Gap1 mean = mean pairwise score gap on Hard test; Gap $>$0.10 = fraction of Hard test pairs exceeding the deployment override threshold. All evaluations use the held-out Ubuntu test set with task-level splits.}
\label{tab:training_ablation}
\begin{tabular}{lcccc}
\toprule
\textbf{Model} & \textbf{Hard} & \textbf{Real Inc.} & \textbf{Gap1 mean} & \textbf{Gap1 $>$0.10} \\
\midrule
 Alignment loss only (baseline) & 0.866 & 0.917 & 0.202 & 69.5\% \\
\quad + Margin ranking loss($m{=}0.10$, $\lambda{=}1.0$) & 0.879 & 0.931 & 0.213 & 71.9\% \\
\quad + Aggressive margin ($m{=}0.20$, $\lambda{=}2.0$) & 0.886 & 0.935 & 0.251 & 76.8\% \\
\quad + MPNet + SigLIP2 + larger model & 0.902 & 0.928 & 0.309 & 79.5\% \\
\quad + Intention-aware action encoder & 0.949 & 0.962 & 0.362 & 87.8\% \\
\quad + Cross-OS pretraining (\ours{}) & \textbf{0.975} & \textbf{0.997} & \textbf{0.373} & \textbf{93.2\%} \\
\bottomrule
\end{tabular}
\end{center}
\end{table}

\begin{wrapfigure}{r}{0.5\textwidth}
  \vspace{-1em}
  \centering
  \captionof{table}{Architecture and input ablation.}
  \label{tab:arch_ablation}
  \small
  \begin{tabular}{lcc}
    \toprule
    \textbf{Model} & \textbf{Hard} & \textbf{Real Inc.} \\
    \midrule
    \multicolumn{3}{l}{\emph{Input ablation}} \\
    \quad Full (baseline) & \textbf{0.866} & \textbf{0.917} \\
    \quad w/o vision & 0.851 & 0.905 \\
    \quad w/o tho.\ \& ref. & 0.786 & 0.829 \\
    \quad w/o code & 0.833 & 0.897 \\
    \quad w/o history & 0.842 & 0.912 \\
    \midrule
    \multicolumn{3}{l}{\emph{Architecture \& paradigm}} \\
    \quad Transformer encoder & 0.755 & 0.769 \\
    \quad Binary classification & 0.073 & 0.117 \\
    \bottomrule
  \end{tabular}

  \vspace{1em}

  \captionof{table}{Ablation on training objectives. CI gap = Correct$-$Incorrect score gap. Alignment loss provides state-action matching structure; margin ranking enforces correct/incorrect separation. Both losses are complementary and necessary.}
  \label{tab:correct_incorrect_gap}
  \scriptsize
  \setlength{\tabcolsep}{3pt}
  \begin{tabular}{lcccc}
    \toprule
    \textbf{Loss} & \textbf{Hard} & \textbf{Real Inc.} & \textbf{Adj gap} & \textbf{CI gap} \\
    \midrule
    Alignment only & 0.866 & 0.917 & 0.202 & 0.005 \\
    Margin only & 0.882 & 0.845 & 0.449 & 0.008 \\
    Both & \textbf{0.886} & \textbf{0.935} & 0.251 & \textbf{0.046} \\
    \bottomrule
  \end{tabular}
  \vspace{-1.3em}
\end{wrapfigure}

\paragraph{Main results.}
Table~\ref{tab:training_ablation} presents the full ablation trajectory on training methodology from the alignment-only baseline to the final \ours{} model. Each row adds one component, isolating its contribution.
\ours{} achieves 97.5\% on adjacent-step discrimination and 99.7\% on real incorrect step detection, which representing a 10.9\% and 8.0\% improvement over the alignment-only baseline, respectively. 
Critically, 93.2\% of Hard test pairs produce score gaps exceeding the 0.10 deployment threshold, meaning the vast majority of \ours{}'s correct decisions would translate to confident overrides at deployment time.
Among individual components, the intention-aware action encoder yields the single largest gain (+4.7\% on Hard), confirming that planning rationale can make the actions more discriminative. Cross-OS pretraining contributes a further +2.6\% despite substantial visual differences between Windows, Mac, and Ubuntu.

Table~\ref{tab:arch_ablation} provides complementary ablations on input modalities and architecture. Removing thought and reflection causes the largest drop ($-$8.0\%), consistent with the intention-aware encoder's importance in Table~\ref{tab:training_ablation}. Code removal ($-$3.3\%) and vision removal ($-$1.5\%) confirm that both executable details and visual context contribute, though less critically. On the architecture side, replacing the dual-encoder with a Transformer encoder degrades performance sharply, and binary classification collapses entirely, validating the contrastive dual-encoder design.



\paragraph{Contrastive alignment vs.\ reward learning.}
\label{sec:alignment_vs_reward}
Table~\ref{tab:correct_incorrect_gap} isolates the effect of each training objective by training with only one loss at a time. Alignment alone reaches Hard accuracy 0.866 but a CI gap of only 0.005: correct and incorrect actions receive nearly identical absolute scores across the dataset. This happens because $\mathcal{L}_{\text{align}}$ optimizes for state-action relevance, and an incorrect adjacent action at step $t{\pm}1$ remains relevant to the same screen as the correct action, so both end up with similar alignment scores. This near-zero gap echoes the collapse problem identified in joint-embedding predictive architectures, where learned encoders can map all inputs to a constant vector that trivially minimizes the contrastive loss~\citep{maes2026leworldmodel}. Margin alone shows the opposite trade-off: within-state separation rises to an Adj gap of 0.449 (the largest of the three), but Real Incorrect detection drops 9.0 points to 0.845. The margin loss enforces only a relative ordering between correct and incorrect at each state and leaves the absolute score range unconstrained, which hurts cross-state discrimination such as Real Incorrect, where the negative is drawn from a different step of the same task. Combining both losses yields the largest CI gap (0.046, roughly 9$\times$ alignment-only) and the best accuracy on both Hard (0.886) and Real Incorrect (0.935). The two losses act on different axes: alignment anchors the embedding space so absolute scores carry consistent meaning across states, and margin ranking sharpens the per-state correctness signal that alignment cannot produce on its own.


\paragraph{Cross-OS pretraining improves target-domain performance.}
Despite substantial visual differences between Windows, Mac, and Ubuntu, pretraining on 398K cross-OS steps before finetuning on 77K Ubuntu steps yields +2.6\% on Hard and +3.5\% on Real Incorrect. This suggests that abstract GUI patterns (dialog flows, menu hierarchies, form layouts) are sufficiently shared across operating systems to transfer reward estimation capability, consistent with the small inter-MDP structural distance discussed in Section~\ref{sec:ProblemFormulation}.

Additional ablations on style robustness, encoder choices, history architectures, and error analysis are provided in Appendix~\ref{app:additional_ablations}.

\subsection{Online evaluation on OSWorld}
\label{sec:deployment}

\begin{table*}[t]
\begin{center}
\scriptsize
\caption{Per-domain success rate (\%) on OSWorld. TDB = Thunderbird. \good{Green} indicates improvement over baseline; \bad{red} indicates regression. \ours{} improves Agent S3 on 7 of 10 domains under the constrained deployment configuration.}
\label{tab:per_domain}
\setlength{\tabcolsep}{4pt}
\begin{tabular}{lcccccccccc|c}
\toprule
Model & Chrome & GIMP & Calc & Impress & Writer & Multi & OS & TDB & VLC & VSCode & \textbf{All} \\

\midrule
S3 (baseline) & 53.2 & 53.8 & 42.6 & 46.8 & 52.2 & 29.0 & 58.3 & 66.7 & 29.4 & 60.9 & 45.2 \\
S3 +\ours{} & 65.2 & 53.8 & 51.1 & 44.7 & 56.5 & 38.7 & 70.8 & 66.7 & 47.1 & 65.2 & \textbf{52.1} \\
\midrule
$\Delta$ & \good{+12.0} & = & \good{+8.5} & \bad{-2.1} & \good{+4.3} & \good{+9.7} & \good{+12.5} & = & \good{+17.7} & \good{+4.3} & \good{+6.9} \\
\bottomrule
\vspace{-0.7cm}
\end{tabular}
\end{center}
\end{table*}

\begin{wrapfigure}{r}{0.52\textwidth}
\vspace{-1.5em}
\centering
\small
\captionof{table}{End-to-end task success rate on OSWorld (361 tasks). SR = success rate, OR = override rate.}
\label{tab:osworld_results}
\begin{tabular}{lccc}
\toprule
\textbf{System} & \textbf{SR} & \textbf{OR} \\
\midrule
 S3 w/ N=3(baseline) & 45.2\%  & --- \\
 S3 + random selection & 44.2\% \down{1.0} & 56.9\% \\
 S3 + LLM self-evaluation & 45.8\% \up{0.6}& 52.1\% \\
 S3 + Alignment only & 48.0\% \up{2.8} & 40.1\% \\
 S3 + \ours{} & \textbf{52.1\%} \up{6.9} & 46.6\% \\
\bottomrule
\end{tabular}
\end{wrapfigure}

\paragraph{Main results.}
Table~\ref{tab:osworld_results} reports end-to-end task success rates on OSWorld. All methods generate $N=3$ candidate actions per step with the same LLM backbone and visual grounding model. We deliberately use a constrained configuration: fewer parallel candidates ($N=3$), shorter horizon (50), a smaller reasoning backbone (GPT-5-mini), and coding-based actions disabled, compared to the SOTA Agent S3 setup. 
This constrained setting both reflects realistic deployment conditions and amplifies the effect of action selection errors: with fewer steps and weaker planning, each individual action matters more, providing a more sensitive testbed for evaluating \ours{}'s scoring capability. Under these conditions, random selection and LLM self-evaluation show no meaningful improvement over the baseline, indicating that naive diversification and self-preference-biased evaluation are insufficient. The alignment-only scorer reaches 48.0\%, showing that learned representations already outperform LLM-based selection. \ours{} achieves 52.1\% (+6.9 over baseline), with a 46.6\% override rate indicating selective rather than indiscriminate intervention.

\paragraph{Per-domain analysis.}
Table~\ref{tab:per_domain} breaks down success rates by application domain. \ours{} improves 7 of 10 domains, with the large gains on Chrome (+12.0), and multi-app tasks (+9.7), i.e., domains where tasks involve navigating nested menus and targeting small UI elements, precisely the setting where distinguishing between similar-looking candidate actions matters most. 
GIMP and Thunderbird show no change, likely within noise given the small per-domain sample sizes. The slight regression on Impress ($-$2.1) may reflect a combination of limited training coverage for presentation-specific operations and the domain's high dependence on precise visual grounding, where the frozen vision encoder's limitations become a bottleneck that the scorer cannot compensate for.

\begin{wrapfigure}{r}{0.52\textwidth}
\centering
\small
\captionof{table}{Scorer behavior across 361 OSWorld tasks.}
\label{tab:scorer_behavior}
\begin{tabular}{lr}
\toprule
\multicolumn{2}{l}{\emph{Step-level decision distribution}} \\
\midrule
Total steps & 6,826 \\
\quad Dedup (all candidates identical) & 2,288 (33.5\%) \\
\quad Scored by model & 4,538 (66.5\%) \\
\midrule
\multicolumn{2}{l}{\emph{Among scored steps}} \\
\midrule
Agree (scorer confirms \#1) & 2,118 (46.7\%) \\
Defer (low confidence) & 304 (6.7\%) \\
Override (scorer selects different) & 2,116 (46.6\%) \\
\midrule
\multicolumn{2}{l}{\emph{Score statistics}} \\
\midrule
Mean selected score & 0.490 \\
Mean top1--top2 gap & 0.118 \\
Mean unique candidates / step & 2.19 \\
\bottomrule
\end{tabular}
\end{wrapfigure}

\paragraph{Scorer behavior analysis.}
Table~\ref{tab:scorer_behavior} breaks down \ours{}'s decisions across 6,826 steps. One-third of steps are deduplicated before scoring, reflecting limited diversity among $N=3$ candidates—the mean unique candidates per step is only 2.19. Among scored steps, the scorer overrides Agent S3's default in 46.6\% of cases, with a mean top-1 to top-2 gap of 0.118, confirming that overrides are backed by meaningful score separation rather than marginal differences. The 6.7\% defer rate shows the threshold $\sigma$ effectively filters out-of-distribution inputs where the scorer lacks reliable signal.

\paragraph{Offline-to-online gap analysis.}
While \ours{} achieves 97.5\% pairwise accuracy offline, the online improvement (+6.9 points) is more modest, reflecting several compounding factors. First, distribution shift between AgentNet training data and Agent S3's outputs remains imperfectly bridged (Appendix~\ref{app:field_mapping}): Agent S3 produces longer, multi-statement executable code blocks and structured planning formats that differ from AgentNet's single-action annotations, creating input patterns unseen during training. Second, the action space at deployment is more diverse than the offline test set: Agent S3 generates free-form candidates whose phrasing, coordinate conventions, and code style vary substantially from AgentNet's annotation schema, reducing scorer confidence on novel inputs. Third, errors in sequential decision-making accumulate: once the scorer selects a suboptimal action, or fails to override an incorrect one, the resulting state diverges from any training trajectory, making subsequent scoring increasingly unreliable. These factors jointly limit the translation from offline discrimination accuracy to online task completion, and suggest that closing this gap requires either training on agent-generated data or online adaptation of the reward model.


\section{Conclusion}
\label{sec:conclusion}

We presented \ours{}, a plan-aware reward model that scores Computer-Use Agent actions by learning from multi-environment offline GUI trajectories. Our results demonstrate that contrastive alignment and reward learning serve complementary roles, that planning intent is highly informative for action evaluation, and that reward estimation capability transfers across operating system boundaries. Deployed as a re-ranker for Agent S3 on OSWorld, an environment entirely unseen during training, \ours{} improves task success rate by 6.9 points with a single 13M-parameter model. The gap between offline discrimination accuracy (97.5\%) and online improvement suggests that closing the distribution shift between training data and deployment agents, and extending reward estimation to multi-step lookahead, are promising directions for future work.


\bibliography{ref}
\bibliographystyle{iclr2026_conference}

\newpage

\appendix
\section{ Model architecture details}
\label{app:architecture}

Table~\ref{tab:architecture} summarizes the detailed specifications of each component in the \ours{} reward model.

\begin{table}[h]
\begin{center}
\scriptsize
\caption{Detailed architecture of \ours{}. All text fields use frozen all-mpnet-base-v2 embeddings (768d); screenshots use frozen SigLIP-SO400M embeddings (1152d). The shared output dimension is 384.}
\label{tab:architecture}
\begin{tabular}{lp{10cm}}
\toprule
\textbf{Component} & \textbf{Specification} \\
\midrule
\textbf{In-context history encoder} & \\
\quad Input per step & $[\mathbf{v}_{t'} \| \text{obs}_{t'} \| \text{act}_{t'} \| \text{code}_{t'} \| \text{xy}_{t'}] \in \mathbb{R}^{3074}$ \\
\quad Step projection & Linear(3074, 512) $\to$ GELU $\to$ LayerNorm(512) \\
\quad Sequence model & 2-layer GRU, hidden size 384, dropout 0.1 \\
\quad Output & Last hidden state $\mathbf{h} \in \mathbb{R}^{384}$ \\
\quad Padding & Zero-padding when fewer than $K{=}3$ previous steps exist \\
\midrule
\textbf{State encoder} & \\
\quad Input & $[\mathbf{v}_t \| \text{obs}_t \| \text{inst} \| \text{reflect}_t \| \mathbf{h}] \in \mathbb{R}^{3840}$ \\
\quad MLP & Linear(3840, 1024) $\to$ GELU $\to$ LN(1024) $\to$ Dropout(0.1) $\to$ Linear(1024, 768) $\to$ GELU $\to$ Linear(768, 384) \\
\quad Output & L2-normalized $\mathbf{s} \in \mathbb{R}^{384}$ \\
\midrule
\textbf{Action encoder} & \\
\quad Input & $[\text{thought}_t \| \text{act}_t \| \text{code}_t \| \text{xy}_t] \in \mathbb{R}^{2306}$ \\
\quad MLP & Linear(2306, 1024) $\to$ GELU $\to$ LN(1024) $\to$ Dropout(0.1) $\to$ Linear(1024, 512) $\to$ GELU $\to$ Linear(512, 384) \\
\quad Output & L2-normalized $\mathbf{a} \in \mathbb{R}^{384}$ \\
\midrule
\textbf{Temperature} & Learnable $\tau$, init $e^{\log 0.07} \approx 0.07$, clamped to $[0.01, 1.0]$~\citep{clip} \\
\midrule
\textbf{Total} & $\sim$13M trainable parameters \\
\bottomrule
\end{tabular}
\end{center}
\end{table}






\section{Data statistics}
\label{app:data_splits}

45.9\% of Ubuntu tasks contain at least one incorrect step, providing hard negatives for the margin ranking loss.
All evaluation uses a \emph{task-level} split of the Ubuntu subset: 85\% train / 10\% validation / 5\% test, ensuring no step from a test task appears during training. The cross-OS data (Windows + Mac) is used only during Stage~1 pretraining and does not overlap with the evaluation split. The effectiveness of cross-OS transfer relies on shared geometric structure across MDP families---an assumption supported by recent theoretical work on structuring value representations via geometric coherence~\citep{zhang2026structuring}.

Ubuntu subset key statistics:
\begin{itemize}
    \item Task completion: True = 2,316 (46.3\%), False = 357 (7.1\%), None = 2,327 (46.5\%)
    \item Steps with incorrect actions: 45.9\% of tasks contain $\geq$1 incorrect step
    \item Thought field present: 59\% of steps (45,750 / 77,448)
    \item XY coordinate range: normalized $[0, 1]$
    \item Screenshot resolution: 1920$\times$1080
    \item Consecutive screenshot overlap: $z_{\text{after}}[i] = z_{\text{before}}[i+1]$ within same task (verified)
\end{itemize}

\begin{table}[h]
\begin{center}
\caption{AgentNet dataset statistics by operating system.}
\begin{tabular}{lccc}
\toprule
& \textbf{Ubuntu} & \textbf{Windows + Mac} & \textbf{Total} \\
\midrule
Tasks & 5,000 & 17,625 & 22,625 \\
Steps & 77,448 & 321,380 & 398,828 \\
Thought present & 59\% & varies & --- \\
\bottomrule
\end{tabular}
\end{center}
\end{table}

\section{Training hyperparameters}
\label{app:hyperparams}

\begin{table}[h]
\begin{center}
\caption{Complete training hyperparameters for \ours{}.}
\begin{tabular}{lll}
\toprule
\textbf{Parameter} & \textbf{Stage 1 (Pretrain)} & \textbf{Stage 2 (Finetune)} \\
\midrule
Data & 398K steps (Ubuntu+Win+Mac) & 77K steps (Ubuntu) \\
Epochs & 30 & 40 \\
Learning rate & $5 \times 10^{-4}$ & $1 \times 10^{-4}$ \\
Margin weight ($\lambda$) & 2.0 & 3.0 \\
Margin ($m$) & 0.20 & 0.20 \\
Batch size & 1024 & 1024 \\
Optimizer & AdamW & AdamW \\
Weight decay & $10^{-4}$ & $10^{-4}$ \\
Scheduler & CosineAnnealingLR & CosineAnnealingLR \\
Min learning rate & $10^{-6}$ & $10^{-6}$ \\
Gradient clipping & max norm 1.0 & max norm 1.0 \\
Temperature init & 0.07 & from Stage 1 \\
Temperature range & $[0.01, 1.0]$ & $[0.01, 1.0]$ \\
History length & 3 steps & 3 steps \\
Embedding dim & 384 & 384 \\
Text encoder & all-mpnet-base-v2 (768d) & same \\
Vision encoder & SigLIP-SO400M (1152d) & same \\
Text augmentation & 75\% prob, 30--50\% dropout & same \\
Augment variants & 3 per step & same \\
Task weights & 1.0 / 0.3 / 0.7 & same \\
Trainable params & $\sim$13M & $\sim$13M \\
Hardware & Single A100 GPU & same \\
\bottomrule
\end{tabular}
\end{center}
\end{table}

\section{Additional ablations}
\label{app:additional_ablations}

\paragraph{Style robustness.}
A critical requirement for cross-environment generalization is robustness to action text phrasing, as different CUA backbones produce different text styles. The binary classification approach exhibits style sensitivity of 0.999, meaning the score is determined by text style rather than state-action compatibility. Our dual-encoder architecture with text augmentation reduces this to 0.031 (32$\times$ improvement), enabling deployment with LLM backbones unseen during training.

\paragraph{History architectures.} We evaluated GRU 3-step, extended 5-step with visual change signals, skip connections, and parallel modality channels in the alignment-only setting. The simple GRU 3-step consistently performs best, indicating the in-context history encoder is not the bottleneck---consistent with findings in hierarchical RL that flat temporal summaries can outperform deeper skill abstractions when the action space is already well-structured~\citep{li2026arise}. \ours{} retains this design.

\paragraph{Vision encoder.} SigLIP vs.\ SigLIP2 produced near-identical alignment-only results (Hard 0.866 vs.\ 0.858). Adjacent screenshot cosine similarity exceeds 0.94 in both, confirming consecutive GUI screenshots are nearly identical in frozen embedding space. This near-invariance of visual embeddings across consecutive steps parallels challenges in energy-based transition modeling, where learned dynamics must capture meaningful state changes despite high input similarity~\citep{fang2026manifold}. Other related works on visual feature quality and robustness include~\citet{zhao2024balf, bellavia2024image, zhao2026advances}.

\paragraph{Text encoder.} MiniLM (384d) vs.\ MPNet (768d) yielded similar alignment-only results; however, \ours{} uses MPNet for its higher-dimensional embeddings (768d vs.\ 384d), which provides additional capacity for the intention-aware action encoder. With the full \ours{} pipeline, MPNet contributes to the +3.6\% gain shown in the ``MPNet + SigLIP2 + larger model'' row of Table~\ref{tab:training_ablation}.

\paragraph{Negative type matters more than quantity.}
Adjacent-step negatives ($t{\pm}1$) are the most effective training signal for Hard test performance, as they require distinguishing temporally close actions that share nearly identical UI context---a challenge shared by offline RL methods that must learn from suboptimal demonstrations without environment interaction~\citep{fang2024learning}. Adding incorrect-step negatives (labeled $r_t=0$) improves detection of actual mistakes but slightly dilutes the adjacent-step gradient. The final model uses a balanced mix of both.

\paragraph{Margin parameters require careful tuning.}
Increasing the margin from $m=0.10$ to $m=0.20$ and the weight from $\lambda=1.0$ to $\lambda=2.0$ yields a 2.0-point gain on Hard accuracy. The sensitivity to margin parameters echoes findings in reward-confidence calibration for policy optimization~\citep{li2026right}. The finetuning stage uses $\lambda=3.0$, further sharpening reward discrimination for the target Ubuntu domain.

\paragraph{Alignment-only failure analysis.} Analysis of 200 failures from the alignment-only model shows a median score gap of $-0.07$ (narrow margin), with failures split evenly between $t{-}1$ and $t{+}1$ negatives. Nearly all share a pattern: two actions that are both \emph{relevant} to the current UI state but differ in temporal correctness. This failure mode is analogous to the challenge of distinguishing structurally similar but semantically different states in graph-based action evaluation~\citep{li2026acdzero}, and is precisely the gap that the margin ranking loss addresses by injecting an explicit correctness signal.

\paragraph{\ours{} residual failures.} With the reduced failure rate ($\sim$2.5\%), remaining errors concentrate on: (1) steps where the planning intention is absent or uninformative (41\% of AgentNet steps lack intention annotations), limiting the intention-aware action encoder's advantage; and (2) genuine data noise where actions have identical text, code, and coordinates. Incorporating explicit intent inference and uncertainty estimation, as explored in human-machine joint planning systems~\citep{fang2026uncertainty}, could help address the missing-intention cases.

\section{Field mapping for Agent S3 deployment}
\label{app:field_mapping}

Agent S3's output format differs from AgentNet's training data, requiring field mapping at inference time. Table~\ref{tab:field_mapping} summarizes the required transformations.

\begin{table}[h]
\begin{center}
\caption{Field mapping from Agent S3's output format to \ours{}'s input format. The planning intention is extracted from each candidate's plan, enabling per-candidate reward estimation.}
\label{tab:field_mapping}
\begin{tabular}{lll}
\toprule
\textbf{Scorer field} & \textbf{Agent S3 source} & \textbf{Parsing} \\
\midrule
observation & Plan (Screenshot Analysis) & Regex extraction \\
thought & Plan (Next Action) & Per-candidate extraction \\
action & NL description from plan\_code & Comment extraction \\
code & exec\_code & Remove import prefix \\
xy & Pixel coords from exec\_code & Normalize by resolution \\
reflection & reflection field & Direct (null $\to$ empty) \\
\bottomrule
\end{tabular}
\end{center}
\end{table}

\section{Binary classification baselines}
\label{app:binary_baselines}

We also explored binary classification, training an MLP to predict whether a (state, action) pair is correct. Table~\ref{tab:binary_baselines} summarizes this progression. Although progressively adding features (V1$\to$V3) improves AUC from 0.770 to 0.887, these gains are misleading: they are evaluated on a \emph{step-level} split where training and test steps can come from the same task. On the stricter \emph{task-level} split used for all \ours{} evaluations, binary classifiers collapse entirely (Hard accuracy 7.3\%, Table~\ref{tab:arch_ablation}), because they memorize task-specific surface patterns rather than learning generalizable state-action relationships. The Transformer variant (V4) fails even on the step-level split (AUC 0.510, chance level). These failures motivated the switch to the contrastive dual-encoder architecture.

\begin{table}[h]
\begin{center}
\caption{Binary classification baselines (step-level split). Accuracy improves with additional features but is evaluated on an easier split; on the task-level split used for \ours{}, binary classification collapses to 7.3\% (Table~\ref{tab:arch_ablation}).}
\label{tab:binary_baselines}
\begin{tabular}{llcc}
\toprule
\textbf{Model} & \textbf{Features} & \textbf{AUC} & \textbf{Top-1/5} \\
\midrule
V1 baseline & vision + obs + act + inst & 0.770 & 56.4\% \\
V2 & + thought + reflection & 0.847 & 66.5\% \\
V3 & + code & 0.887 & 69.3\% \\
V4 (Transformer) & same as V3 & 0.510 & 29.0\% \\
\bottomrule
\end{tabular}
\end{center}
\end{table}

\section{Case studies}
\label{app:case_studies}

\begin{figure}[ht]
\begin{center}
\includegraphics[width=0.75\linewidth]{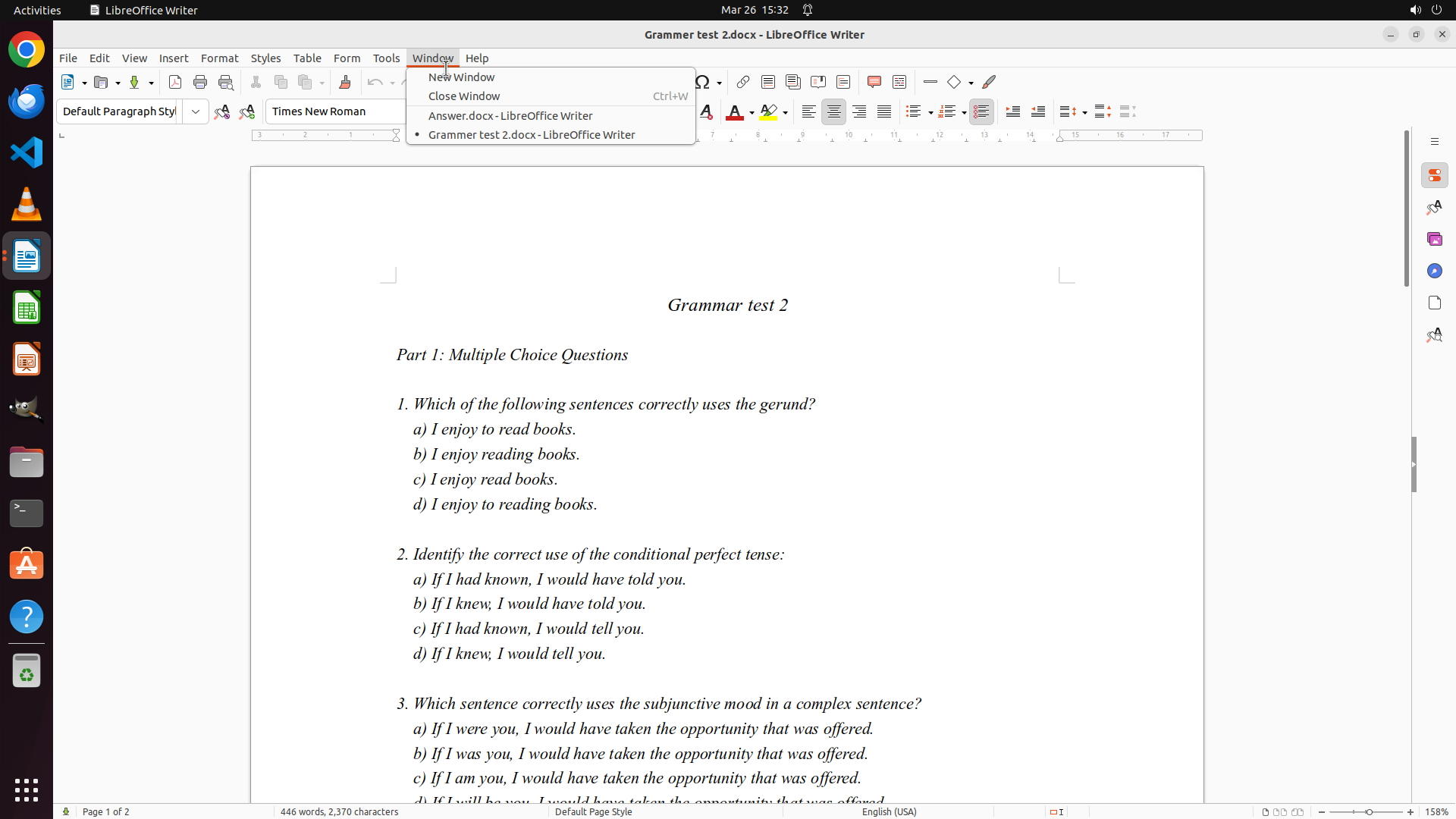}
\vspace{0.3em}

\small
\begin{tabular}{clc}
\toprule
\textbf{\#} & \textbf{Action (intent summary)} & \textbf{Score} \\
\midrule
1 & \texttt{Alt+Tab} \scriptsize{(``use hotkey to switch'')} & 0.401 \\
\textbf{2} & \textbf{Click ``Window'' menu} \scriptsize{(``use menu to select Answer.docx'')} & \textbf{0.637} \\
3 & \texttt{PageDown} \scriptsize{(``scroll to see more questions'')} & 0.308 \\
\bottomrule
\end{tabular}
\end{center}
\caption{Case study 1 (step~7): \ours{} overrides \texttt{Alt+Tab} in favor of the ``Window'' menu for reliable document switching. The score gap of 0.236 reflects the intent-aware encoder's ability to distinguish navigation strategies.}
\label{fig:case_study_nav}
\end{figure}

We present a detailed analysis of a successful OSWorld task (``write grammar test answers into a document'') where \ours{} overrides Agent S3's default candidate at critical steps. The task involves a multi-application workflow: opening grammar test documents, reading questions, switching between windows, typing answers, and saving. Figure~\ref{fig:timeline} visualizes \ours{}'s decisions across all 28 steps; Figures~\ref{fig:case_study_nav}--\ref{fig:case_study_type} detail four representative overrides.

\paragraph{Trajectory overview.}
Across 28 steps, \ours{} overrides Agent S3's default in 10 steps (35.7\%), agrees in 11 steps (39.3\%), and defers to deduplication in 3 steps (10.7\%). Four steps (18--19, 26, 28) generated only a single candidate (no alternatives to compare), so the scorer was not invoked. Override steps with large gaps (e.g., step~7, gap=0.236) reflect high-confidence corrections, while small-gap overrides (e.g., step~27, gap=0.038) can still be critical for task success. Notably, agree steps can also have large gaps (e.g., step~9, gap=0.236), indicating selective intervention rather than indiscriminate overriding.

\begin{figure}[ht]
\centering
\begin{tikzpicture}
\begin{axis}[
    ybar=0pt,
    width=\textwidth,
    height=5cm,
    bar width=7pt,
    ylabel={Top-1 $-$ Top-2 score gap},
    ymin=0, ymax=0.30,
    xtick={1,2,3,4,5,6,7,8,9,10,11,12,13,14,15,16,17,18,19,20,21,22,23,24,25,26,27,28},
    xticklabels={1,2,3,4,5,6,7,8,9,10,11,12,13,14,15,16,17,18,19,20,21,22,23,24,25,26,27,28},
    xticklabel style={font=\tiny},
    xmin=0, xmax=29,
    legend style={at={(0.01,0.97)}, anchor=north west, font=\scriptsize, draw=gray!50, fill=white, fill opacity=0.9, text opacity=1},
    tick label style={font=\scriptsize},
    label style={font=\small},
    ymajorgrids=true,
    grid style={dashed, gray!25},
    axis x line*=bottom,
    axis y line*=left,
    clip=false,
]

\addplot[fill=orange!70, draw=orange!90!black, line width=0.3pt, bar shift=0pt] coordinates {
    (1,0.086) (3,0.001) (7,0.236) (10,0.196) (14,0.021) (16,0.069) (17,0.015) (22,0.005) (24,0.099) (27,0.038)
};

\addplot[fill=teal!50, draw=teal!80!black, line width=0.3pt, bar shift=0pt] coordinates {
    (2,0.136) (6,0.032) (9,0.236) (11,0.024) (12,0.041) (13,0.129) (15,0.206) (20,0.067) (21,0.060) (23,0.136) (25,0.027)
};

\addplot[fill=gray!35, draw=gray!55, line width=0.3pt, bar shift=0pt] coordinates {
    (4,0.008) (5,0.008) (8,0.008)
};

\addplot[fill=white, draw=black!40, line width=0.3pt, bar shift=0pt, postaction={pattern=north east lines, pattern color=black!30}] coordinates {
    (18,0.008) (19,0.008) (26,0.008) (28,0.008)
};

\legend{Override, Agree, Dedup, Single cand.}

\node[above=1pt, font=\small\bfseries, text=red!70!black] at (axis cs:7,0.236) {\ding{72}};
\node[above=1pt, font=\small\bfseries, text=red!70!black] at (axis cs:10,0.196) {\ding{72}};
\node[above=1pt, font=\small\bfseries, text=red!70!black] at (axis cs:16,0.069) {\ding{72}};
\node[above=1pt, font=\small\bfseries, text=red!70!black] at (axis cs:17,0.015) {\ding{72}};

\draw[decorate, decoration={brace, mirror, amplitude=4pt}, thick, gray!70]
    (axis cs:1,-0.055) -- (axis cs:6,-0.055)
    node[midway, below=5pt, font=\tiny, text=gray!80!black] {Open docs};
\draw[decorate, decoration={brace, mirror, amplitude=4pt}, thick, gray!70]
    (axis cs:7,-0.055) -- (axis cs:19,-0.055)
    node[midway, below=5pt, font=\tiny, text=gray!80!black] {Read test 2 \& write answers};
\draw[decorate, decoration={brace, mirror, amplitude=4pt}, thick, gray!70]
    (axis cs:20,-0.055) -- (axis cs:28,-0.055)
    node[midway, below=5pt, font=\tiny, text=gray!80!black] {Read test 3, write answers \& save};

\end{axis}
\end{tikzpicture}
\caption{Decision timeline for a complete OSWorld trajectory (28 steps, task: ``write grammar test answers''). Each bar shows the score gap between the top-ranked and second-ranked candidate, colored by \ours{}'s decision: \textcolor{orange!80!black}{\textbf{override}} (select a different candidate), \textcolor{teal!70!black}{\textbf{agree}} (confirm Agent S3's default), or \textcolor{gray!60}{\textbf{dedup}} (all candidates identical). Red stars mark the four detailed case studies below.}
\label{fig:timeline}
\end{figure}

\paragraph{Case 1: Reliable window navigation via planning intent (step~7).}
The agent has just opened ``Grammer test 2.docx'' and needs to switch back to ``Answer.docx'' to enter the answers. Agent S3's default candidate proposes \texttt{Alt+Tab} (score 0.401), a keyboard shortcut whose target window depends on the OS window stacking order and is therefore unreliable in a multi-document workflow. A third candidate proposes \texttt{PageDown} (score 0.308), which would scroll within the current document rather than switching windows, a clearly incorrect action. \ours{} instead selects Candidate~\#2 (score \textbf{0.637}), which clicks the ``Window'' menu in LibreOffice Writer's menu bar to deterministically select ``Answer.docx'' from the list of open documents. The score gap of 0.236 is the largest in this trajectory, and the discrimination is driven by the planning intent: Candidate~\#2's rationale explicitly states ``I'll switch back to the Answer.docx window using LibreOffice Writer's Window menu instead of using Alt+Tab,'' while Candidate~\#1's rationale mentions ``use a window-switch hotkey.'' The intent-aware action encoder maps these different rationales to distinct embeddings, enabling \ours{} to favor the more reliable navigation strategy.

\paragraph{Case 2: Consistent preference for deterministic navigation (step~10).}
Three steps later, the agent is back in ``Answer.docx'' and needs to switch to ``Grammer test 2.docx'' to read its questions. The Window menu is already open, listing both documents. Agent S3's default proposes \texttt{Ctrl+F6} (score 0.138), a LibreOffice-specific hotkey that cycles through open documents in an order the agent cannot predict. \ours{} overrides to Candidate~\#2 (score \textbf{0.333}), which clicks the ``Window'' menu entry for ``Grammer test 2.docx'' directly. The gap of 0.196 is the second largest in this trajectory. Combined with step~7, this shows that \ours{} \emph{consistently} penalizes non-deterministic navigation hotkeys in favor of explicit menu selection---a pattern learned from offline trajectories where hotkey-based switching frequently leads to wrong-window errors.

\begin{figure}[ht]
\begin{center}
\includegraphics[width=0.75\linewidth]{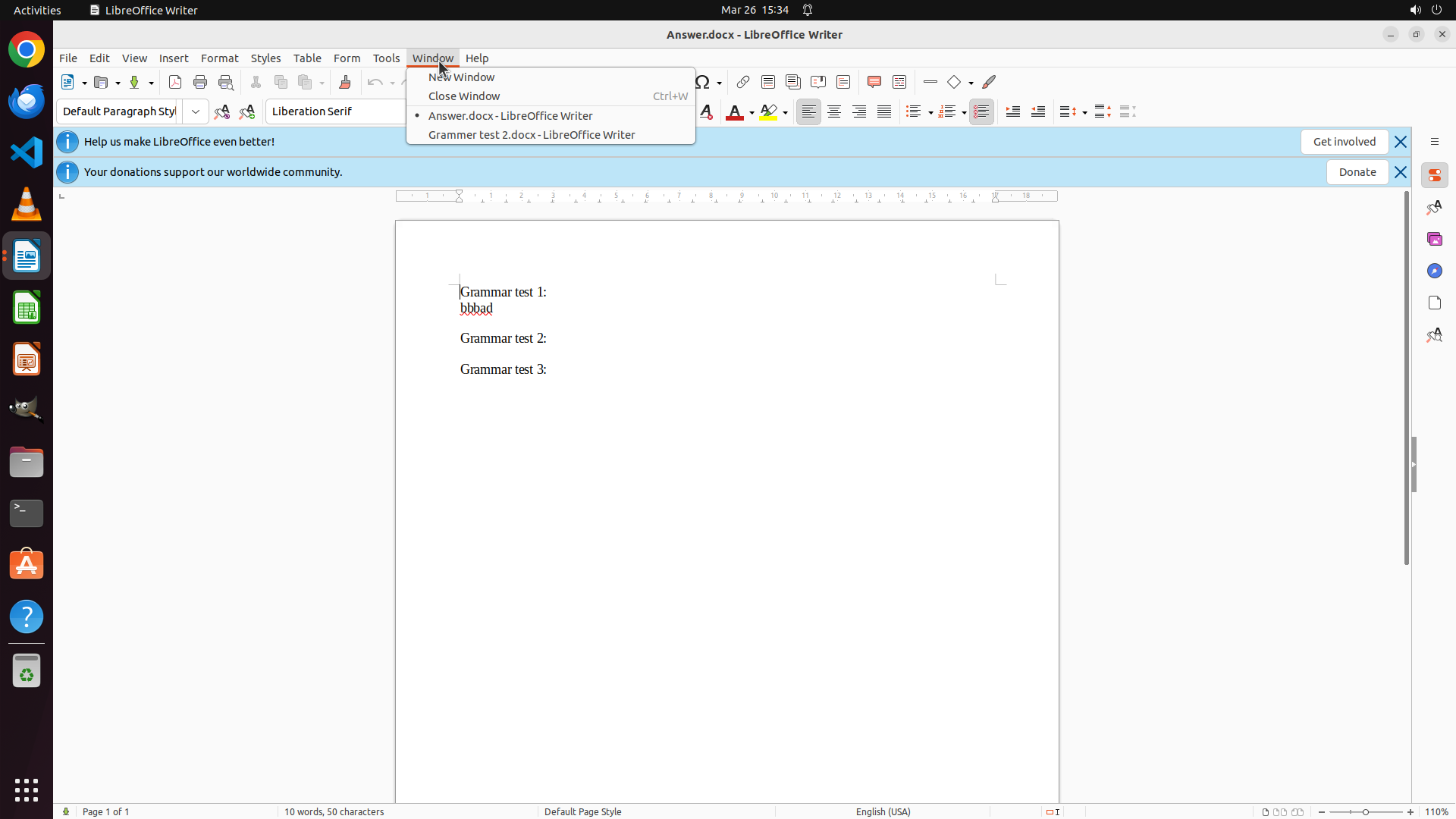}
\vspace{0.3em}

\small
\begin{tabular}{clc}
\toprule
\textbf{\#} & \textbf{Action (intent summary)} & \textbf{Score} \\
\midrule
1 & \texttt{Ctrl+F6} \scriptsize{(``use hotkey to cycle documents'')} & 0.138 \\
\textbf{2} & \textbf{Click ``Window'' $\to$ Grammer test 2} \scriptsize{(``select target from menu'')} & \textbf{0.333} \\
\bottomrule
\end{tabular}
\end{center}
\caption{Case study 2 (step~10): \ours{} again overrides a navigation hotkey (\texttt{Ctrl+F6}) in favor of the ``Window'' menu. The screenshot shows Answer.docx with the Window menu open, listing both documents.}
\label{fig:case_study_nav2}
\end{figure}

\paragraph{Case 3: Precise cursor placement (step~16).}
The agent has returned to ``Answer.docx'' and needs to place the cursor at the correct position to type the Test~2 answers. All three candidates click on the same line ($y{=}0.417$), but at different horizontal positions with different intents: Candidate~\#1 clicks ``in the blank area immediately below `Grammar test 2:'\,'' ($x{=}0.347$, score 0.472); Candidate~\#2 clicks ``on the blank line immediately below the heading'' ($x{=}0.411$, score 0.549); Candidate~\#3 clicks ``immediately to the right of the text `Grammar test 2:'\,'' ($x{=}0.383$, score \textbf{0.618}). \ours{} selects Candidate~\#3, which places the cursor directly after the colon---the most precise insertion point for appending the answer string on a new line. The other two candidates risk placing the cursor on the wrong line or at an offset position, which could corrupt the document formatting. This case demonstrates that the intent-aware encoder distinguishes candidates with nearly identical coordinates but different spatial reasoning.

\begin{figure}[ht]
\begin{center}
\includegraphics[width=0.75\linewidth]{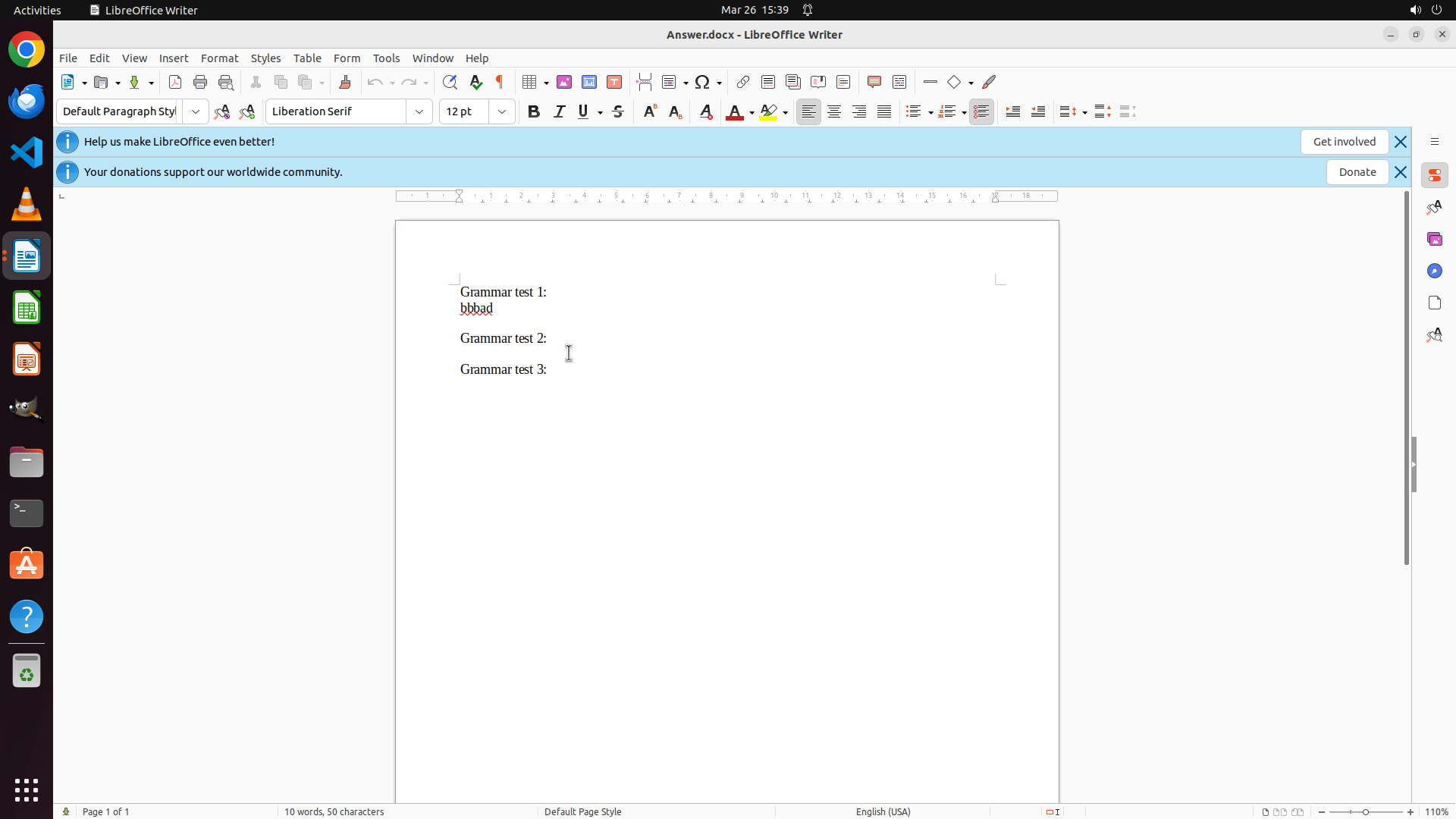}
\vspace{0.3em}

\small
\begin{tabular}{clc}
\toprule
\textbf{\#} & \textbf{Action (intent summary)} & \textbf{Score} \\
\midrule
1 & Click below ``Grammar test 2:'' \scriptsize{(``blank area below'')} & 0.472 \\
2 & Click below the heading \scriptsize{(``blank line below heading'')} & 0.549 \\
\textbf{3} & \textbf{Click right of ``Grammar test 2:''} \scriptsize{(``place cursor after colon'')} & \textbf{0.618} \\
\bottomrule
\end{tabular}
\end{center}
\caption{Case study 3 (step~16): Three click candidates target the same line at slightly different positions. \ours{} selects the one that places the cursor directly after ``Grammar test 2:'' for precise text insertion.}
\label{fig:case_study_cursor}
\end{figure}

\paragraph{Case 4: Format-preserving text entry (step~17).}
Immediately after cursor placement, the agent must type the Test~2 answer string. All three candidates type the same content (``baaad''), but differ in formatting: Candidate~\#1 types \texttt{"baaad"} without a leading newline; Candidate~\#2 types \texttt{"\textbackslash nbaaad"} with a leading newline; Candidate~\#3 types \texttt{"\textbackslash nbaaad"} with an explicit \texttt{enter=False} flag. \ours{} selects Candidate~\#2 (score \textbf{0.494}), which includes the newline character \texttt{\textbackslash n} to place the answer on a new line below the heading---matching the format established for Test~1, where ``bbbad'' appears on its own line beneath ``Grammar test 1:''. Without the leading newline, the answer would be appended directly after the colon on the same line (``Grammar test 2: baaad''), breaking the expected format. Since OSWorld's evaluation scripts verify the final document content against a ground-truth template, incorrect formatting would cause the task to \emph{fail} despite the answer itself being correct. This case shows that \ours{} captures formatting conventions from offline trajectories---a subtle but task-critical signal that distinguishes candidates with identical semantic content.

\begin{center}
\includegraphics[width=0.75\linewidth]{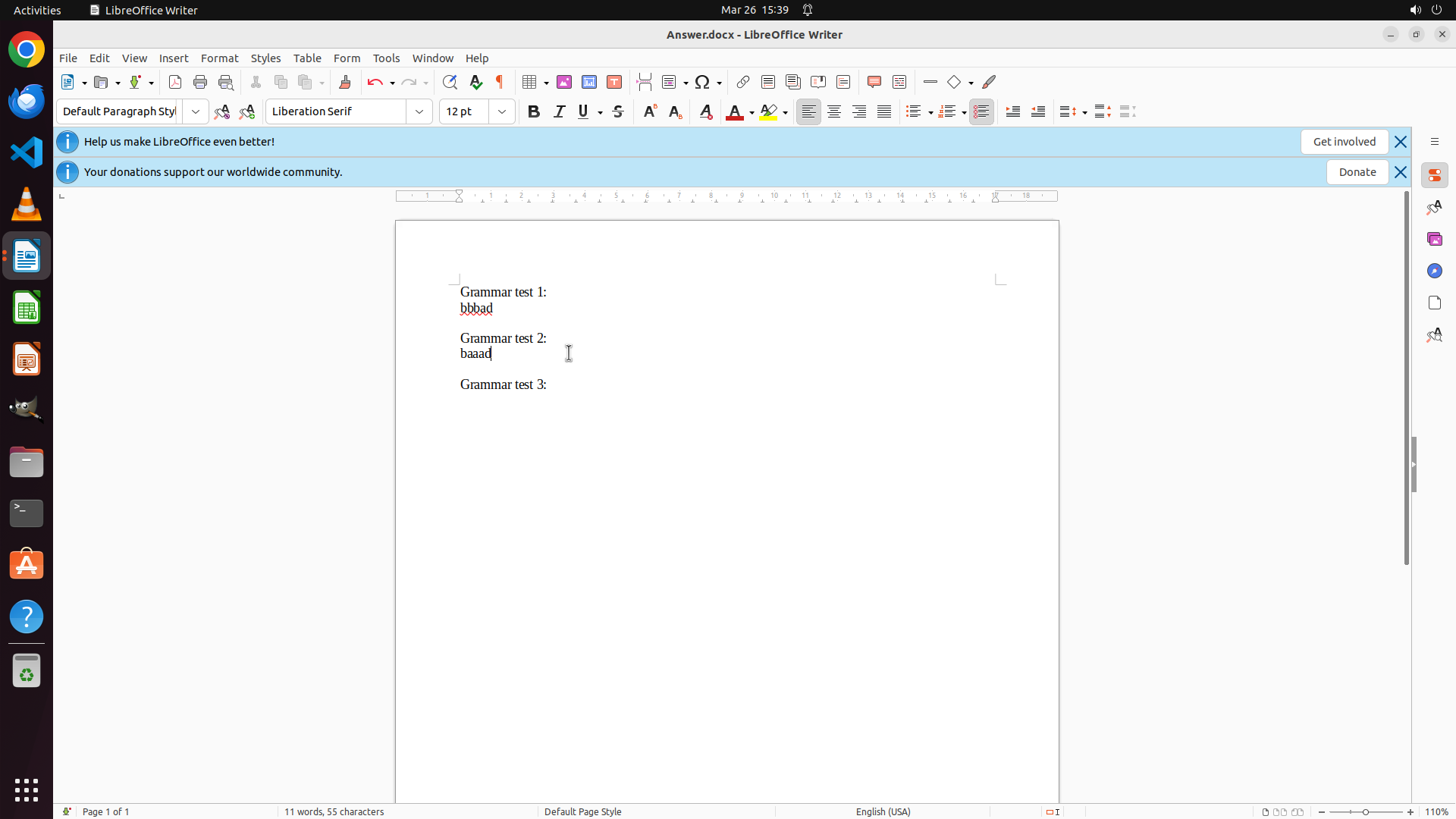}
\vspace{0.3em}

\small
\begin{tabular}{clc}
\toprule
\textbf{\#} & \textbf{Action (intent summary)} & \textbf{Score} \\
\midrule
1 & \texttt{type("baaad")} \scriptsize{(no newline, appends on same line)} & 0.479 \\
\textbf{2} & \textbf{\texttt{type("\textbackslash nbaaad")}} \scriptsize{(newline first, matches Test 1 format)} & \textbf{0.494} \\
3 & \texttt{type("\textbackslash nbaaad", enter=False)} \scriptsize{(newline, explicit no-enter)} & 0.472 \\
\bottomrule
\end{tabular}
\captionof{figure}{Case study 4 (step~17): Three type candidates with identical content but different formatting. \ours{} selects the one with a leading \texttt{\textbackslash n}, preserving the line-by-line format established for Test~1. Without the newline, the answer would appear on the same line as the heading, failing the evaluation.}
\label{fig:case_study_type}
\end{center}

\end{document}